\documentclass[preprint,12pt]{elsarticle}

\usepackage{amssymb}
\usepackage{amsmath}
\usepackage{subfigure}
\usepackage{url}
\usepackage{multirow}

\journal{Nuclear Physics B}

\begin{document}

\begin{frontmatter}



\title{Emotional Cognitive Modeling Framework with Desire-Driven Objective Optimization for LLM-empowered Agent in Social Simulation}




\author[aff1]{Qun Ma}
\ead{1023244018@tju.edu.cn}
\author[aff1]{Xiao Xue\corref{cor1}}
\ead{jzxuexiao@tju.edu.cn}
\author[aff1]{Xuwen Zhang}
\ead{1023244032@tju.edu.cn}
\author[aff1]{Zihan Zhao}
\ead{zhaozihan@tju.edu.cn}
\author[aff1]{Yuwei Guo}
\ead{2024244171@tju.edu.cn}
\author[aff2]{Ming Zhang}
\ead{zhangming1015518539@outlook.com}

\cortext[cor1]{
     Corresponding author. \textit{E-mail address:} \url{jzxuexiao@tju.edu.cn} (Xiao Xue).  
}

\affiliation[aff1]{organization={College of Intelligence and Computing},
            addressline={Tianjin University}, 
            city={Tianjin},
            postcode={300350},
            country={China}}

\affiliation[aff2]{organization={Faculty of Environment, Science and Economy},
            addressline={University of Exeter}, 
            city={Exeter},
            postcode={EX4 4QJ},
            country={UK}}

\begin{abstract}
The advent of large language models (LLMs) has enabled agents to represent virtual humans in societal simulations, facilitating diverse interactions within complex social systems. However, existing LLM-based agents exhibit severe limitations in affective cognition: They fail to simulate the bounded rationality essential for bridging virtual and real-world services; They lack empirically validated integration mechanisms embedding emotions within agent decision architectures. This paper constructs an emotional cognition framework incorporating desire generation and objective management, designed to achieve emotion alignment between LLM-based agents and humans, modeling the complete decision-making process of LLM-based agents, encompassing state evolution, desire generation, objective optimization, decision generation, and action execution. This study implements the proposed framework within our proprietary multi-agent interaction environment. Experimental results demonstrate that agents governed by our framework not only exhibit behaviors congruent with their emotional states but also, in comparative assessments against other agent types, demonstrate superior ecological validity and generate decision outcomes that significantly more closely approximate human behavioral patterns.
\end{abstract}



\begin{keyword}
LLM-based agent \sep cognitive modeling \sep emotion alignment \sep Computational Experiment


\end{keyword}

\end{frontmatter}


\section{Introduction}
In the field of LLM-based agent simulation, the significance of emotional cognition is gaining increasing recognition. This research direction not only provides profound insights into understanding complex human emotional and cognitive processes but also encompasses multidimensional applications. These include analyzing the emotional states of individuals or groups, as well as effectively leveraging these emotional insights in scenarios such as social media analytics, human-computer interaction (HCI), and mental health assessment. LLM-based agents endowed with emotional cognition capabilities can more closely align with human values, thereby significantly enhancing their performance in emotion-related downstream tasks. In addition, LLM-based agent simulation requires some core characteristics, especially \textbf{Bounded Rationality}, achieved through constrained information processing and anthropomorphic heuristic decision-making, balancing efficiency and rationality in complex interactions to enable credible human-AI symbiosis \citep{dang2025dynamic}.

\begin{figure}[h]
    \centering
    \includegraphics[height=4cm, width=0.9\textwidth]{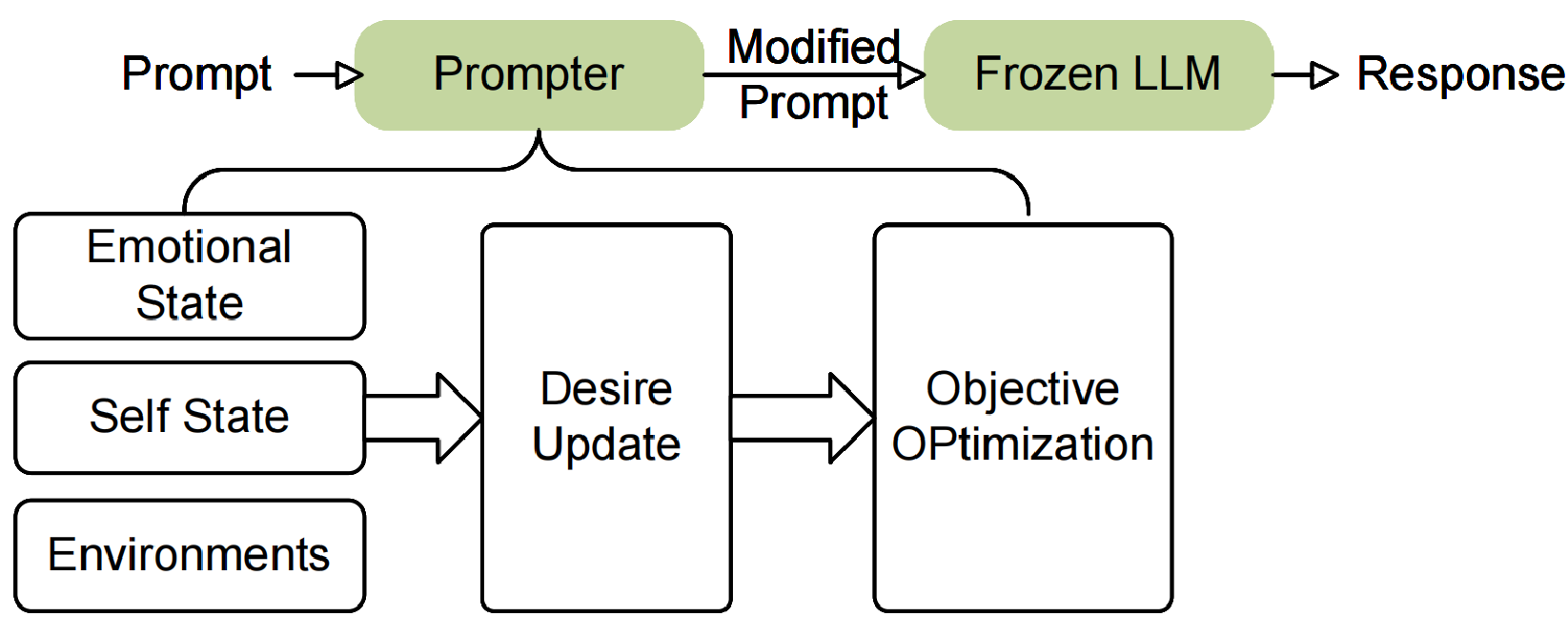}
    \caption{Illustration of embedding emotion into decision core of LLM-based agents.}
    \label{fig:enter-label}
\end{figure}

Current research on emotional cognition in LLMs and LLM-based agents predominantly focuses on the processing and analysis of emotinoal states, encompassing emotion recognition \citep{zhang2024refashioning}, emotional response generation \citep{chen2024temporalmed}, and the evaluation of emotional cognition capabilities \citep{sap2022neural}. By designing and conducting extensive computational experiments in the virtual world, researchers can identify potential social risks and governance crises related to technological governance while effectively constructing social systems \cite{xue2018social, 9772408, 10.1145/3686802}. Researchers enhance the emotional processing abilities of LLMs and LLM-based agents through methods such as fine-tuning and in-context learning. However, studies on such "emotionally intelligent assistants" fall short of meeting the requirements for LLM-based agent simulation. This is because emotional cognition extends beyond initial recognition and comprehension to include subsequent interactive feedback loops within the agent's behavioral processes. Research \citep{adolphs2012interaction} indicates that emotion serves as an indispensable adaptive mechanism for managing complex organizational systems, exerting significant influence on memory, attention, and reasoning. In systems equipped with reasoning mechanisms yet devoid of emotional factors, decisions and judgments—despite the system remaining fully operational—prove suboptimal.

The emotional cognition framework proposed in this study facilitates interaction and feedback loops between an agent's emotions and its subsequent decision-making processes, building upon the foundation of robust state perception within LLM-based agents. This approach not only enables LLM-based agents to establish an interpretable "state-decision-action" cycle but also enhances the authenticity and trustworthiness of LLM-based agent simulation. However, the implementation of this process presents a significant challenge (Figure 1): How emotions affect LLM-based agents' behaviors or decisions. Since LLMs govern the decision generation of LLM-based agents, it remains challenging for researchers to embed emotions into this generative process. Martinez et al. \citep{martinez2005emotions} posit that emotion's tighter coupling with goals (compared to specific behavioral responses) explains why it leads to richer, more variable, and more flexible behaviors. Consequently, fluctuations in emotional states within LLM-based agents trigger distinct motivational drivers-desire, which in turn elicit novel objective formation. This mechanism ultimately enables the emergence of naturalistic human-like behavioral patterns.

In this study, we construct an emotional cognition framework incorporating desire generation and objective management, designed to achieve emotion alignment between LLM-based agents and humans. Distinct from traditional fine-tuning or in-context learning, this framework models the complete decision-making process of LLM-based agents, encompassing state evolution, desire generation, objective optimization, decision generation, and action execution. This enables agents to dynamically generate distinct desires in response to state changes and consequently update their planning objectives, thereby guiding them to proactively make emotion-driven decisions. This study implements the proposed framework within our proprietary multi-agent interaction environment. The environment not only replicates activity scenarios analogous to real-world societies but also facilitates combative simulations between diverse agent types, including agents driven by rule-based decision-making, reinforcement learning (RL)-based decision-making, and GPT-based decision-making. Experimental results demonstrate that agents governed by our framework not only exhibit behaviors congruent with their emotional states but also, in comparative assessments against other agent types, demonstrate superior ecological validity and generate decision outcomes that significantly more closely approximate human behavioral patterns.

Our main contributions can be summarized as follows:
\begin{itemize}
\item \textbf{A Cognitive Modeling Framework for Emotion Alignment in LLM-Based Agents.} This novel framework establishes influence pathways through which emotional states impact agent decision-making. By leveraging state variations to propel desire generation and incentivize objective optimization, it enables holistic emotional cognition within LLM-based agents.
\item \textbf{Multi-Agent Interaction Scenario Simulator.} We developed a computational experiment-ready multi-agent scenario simulator based on the Repast4py framework. This simulator provides robust experimental design and analytics capabilities, facilitating further exploration of human-like activity generation.
\item \textbf{Experiments.} Through comprehensive comparative analysis, we demonstrate that agents governed by this framework consistently exhibit behaviors exhibiting high congruence with their intrinsic emotional states. Furthermore, their constituent multi-agent social systems manifest emergence phenomena that more closely align with real-world social systems. Crucially, when compared to other agent types, these agents generate activity sequences that exhibit the bounded rationality essential for LLM-based agent simulation, thereby validating both the framework's efficacy and the robustness of its social simulations.
\end{itemize}

\section{Related Work}
\textbf{Emotional Cognition for LLMs and LLM-based Agent.} Emotional cognition capabilities enable bounded rationality in LLM-based agents and enhance the authenticity of their simulations. However, existing research predominantly targets performance enhancement in application scenarios, focusing primarily on LLMs' processing and analysis of emotions. 1) Emotion Recognition \& Classification:
Ratican and Hutson proposed a six-dimensional emotional model analyzing emotions across dimensions including arousal level, valence, dominance, agency, affiliation, and novelty \cite{ratican2023six}. Rodríguez-Ibáñez et al. evaluated emotional analysis methodologies in social networks and their applications in domains such as stock market valuation, political analysis, and cyberbullying prevention education \cite{rodriguez2023review}. 2) Generative Task Evaluation:
Xie et al. investigated emotional generation capacities in LLMs, focusing on stylistic, register, and length variations in story creation, revealing LLMs' significant advantages in narrative content generation \cite{xie2023next}. Zhong et al. introduced a dynamic memory mechanism enabling LLMs to leverage historical emotional interactions for current decision-making through cross-temporal contextualization \cite{zhong2024memorybank}. 3) Human Mental State Inference \& Replication:
Zhu et al. examined LLMs' capabilities in psychological inference tasks, particularly in deducing users' latent goals and fundamental psychological needs \cite{zhu2024reading}. Xu et al. assessed LLM performance in mental health prediction tasks, highlighting the necessity for bias mitigation interventions \cite{xu2307leveraging}.

\textbf{Emotional Decisions:}   
Emotion plays a pivotal role in individual decision-making processes, significantly modulating cognition and reasoning. Substantial research focuses on embedding emotional components into agents' decision-making workflows during simulations: 1) Direct Modulation of Reasoning Processes: Emotional states introduce additional evaluative dimensions (e.g., certainty, risk perception) into core reasoning mechanisms, thereby modifying deliberative content \cite{marques2025predicting}. Concurrently, they reshape cognitive schemas toward reasoning subjects, effectively regulating reasoning depth \cite{zhang2024simplicity}. 2) Indirect Goal Activation Priming:
Emotions exhibit tight coupling with motivational drivers, where emergent desires catalyze subsequent goal optimization – initiating novel reasoning cycles \cite{frijda2017laws}. Nevertheless, research on this emotionally-triggered decision paradigm remains underdeveloped. 3) Post-Hoc Evaluative Intervention:
This approach quantitatively assesses consequential emotioal state trajectories to enable real-time decision recalibration \cite{huang2023emotionally}.

\begin{figure*}[t!]
    \centering
    \includegraphics[width=\textwidth]{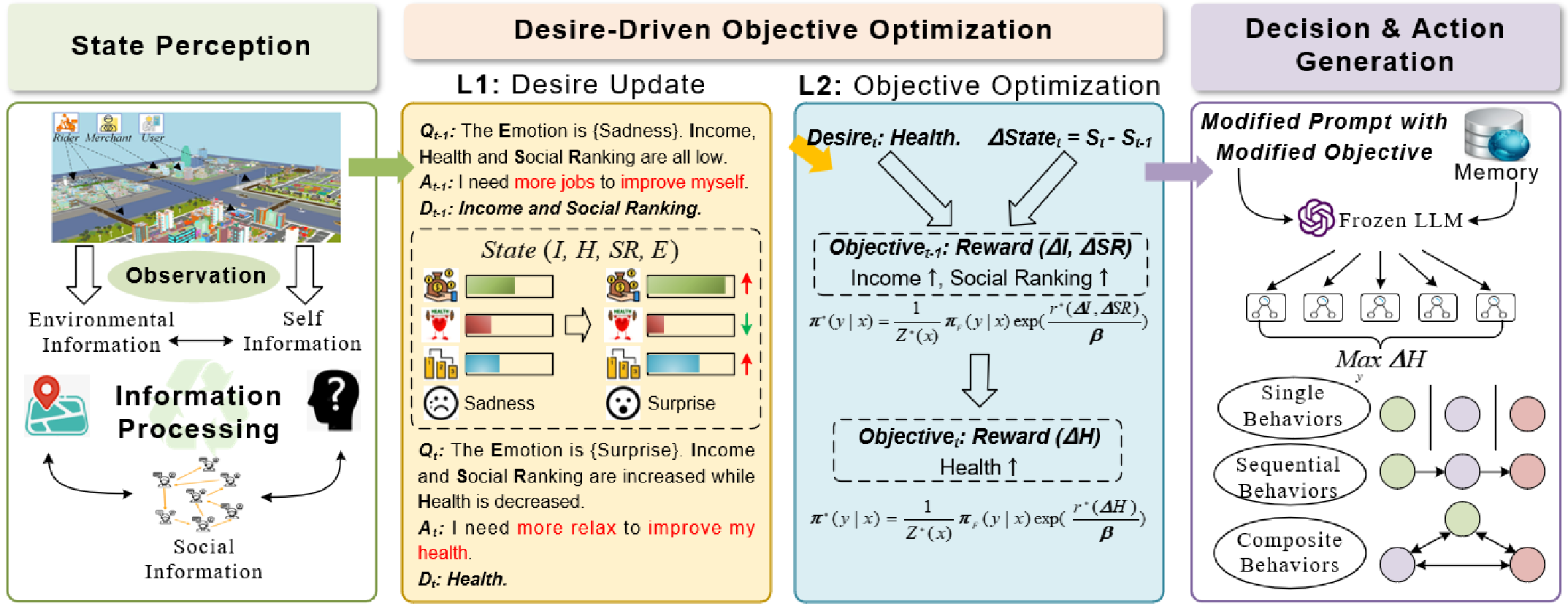}
    \caption{Our Emotional Cognitive Modeling Framework for LLM-based agents.}
\end{figure*}

\section{Emotional Cognitive Modeling Framework}
We formulate the problem of achieving emotion alignment for LLM-based agents in societal simulation, and subsequently present the emotional cognitive modeling framework illustrated in Figure 2.
\subsection{Problem Definition}
When employing LLM-based agents for societal simulation, explicitly defining and resolving challenges related to behavioral alignment and simulation fidelity is imperative. The core objective is to ensure these agents not only generate decisions consistent with their internal states and environmental contexts but also exhibit behavioral patterns approximating human counterparts. Crucially, as emotion serves as a critical modulator in human decision-making, its integration into LLM-based agents is essential for achieving the bounded rationality required in societal simulation. Unlike prevailing LLM-based emotional analysis research, neither fine-tuning enables agents to dynamically respond to real-time emotional state fluctuations in simulations, nor does singular prompt engineering prevent systematic inaccuracies and emotional hallucinations in emotion-influenced decisions. Consequently, this study addresses the fundamental question: How to embed emotional cognition within the decision-making modules of LLM-based agents?
\subsection{Overview}
Emotional Cognitive Modeling Framework comprises three core modules: \textit{Information Processing System}, \textit{Desire-Driven Objective Optimizer}, and \textit{Decision-Behavior System}, augmented with essential agent components including memory and self-reflection mechanisms. Information Processing System performs integration of environmental data, internal states, and social information. Desire-Driven Objective Optimizer generates novel desire vectors $D_t$ based on state representations $State_t$ received from Information Processing System. This drives iterative objective refinement $Objective_t$ and constructs dynamic reasoning prompts $Prompt_t$. Decision-Behavior System executes reasoning-based planning using $Objective_t$-conditioned $Prompt_t$, ultimately selecting behavioral outputs.

\subsection{Step 1: State Perception}

To simulate human behaviors, this study constructs an Information Processing System within agents to perceive state transitions in simulated environments. Leveraging our proprietary multi-agent interaction framework, we formally define the composite state representation $State_t(I_t, H_t, SR_t, E_t)$, where $I_t$ denotes agent income at time $t$, $H_t$ represents health status (simplified as stamina points), $SR_t$ indicates social rank (simplified as income percentile ranking), $E_t$ captures the emotional state
(Environmental parameters will be detailed in the experimental section). The subsequent section elaborates the formal modeling methodology for $E_t$.

Inspired by Retrieval-Augmented Generation (RAG), this study integrates the EDBE Sentiment Dataset \cite{sun2023new} as an external knowledge repository for emotional decision information retrieval. There are 7 types of emotional features in the given emotional dataset, including happiness, anger, disgust, surprise, fear, sadness and neutral, constructed using the PAD value for LLM-based agents \cite{wen2021automatically}. The PAD value composed of three dimensions as the followings. 

Pleasure: embodies both the positive (positive) and negative (negative) aspects of the user's emotional state, which 
can be seen as proportional to the change in their earnings: $Pleasure_t = k_p(I_t-I_{t-1})$.

Arousal: reflects the activation level of the user's nerves (physiological level) and the degree of arousal, which
can be seen as proportional to the change in its energy value: $Arousal_t = k_a(H_t-H_{t-1})$.

Dominance: reflects the strength of mutual dominance between the user and the external environment (user dominance is positive, external dominance is negative), which
scales with the revenue tier of agents within the ecosystem: $Dominance_t = criteria[SR_t]$.

\subsection{Step 2: Desire-Driven Objective Optimization}
\textbf{L1: Desire Update.}
Theory of Needs \cite{maslow1943theory} posits that individuals' behavioral objectives originate from fulfilling specific needs or desires (e.g., social status, self-worth). Furthermore, humans dynamically prioritize these needs based on unique attributes—particularly emotional states. These factors highlight the complexity of human motivation while elucidating how demands drive behavioral patterns.

This paper establishes a triadic state attribute system (income, health status, social rank) as the agent's desire architecture, with emotional state $E_t$ serving as the sole triggering mechanism for desire updating. For instance: 1) When an agent primarily driven by income goals experiences significant emotional transition (sadness → surprise) at timestep $t$; 2) The desire updating process initiates and detects abnormal health status $H_t$ while $I_t$ and $SR_t$ remain stable or improve; 3) The new desire vector $D_t$ updates to prioritize $H_t$ enhancement.

\textbf{L2: Objective Optimization.}
In societal simulation, the decision-making process of LLM-based agents constitutes an inherent optimization problem, where the objective function—as its core conceptual element—drives agents to seek optimal solutions under constraints. The current phase aims to generate modified prompts incorporating the renewed objective function $Objective_t$, derived from the updated desire vector $D_t$ (provided by Module L1) and state variations $\Delta S$.

For a given prompt $x$, the probability of generating a response $y$ from the frozen model is represented by $\pi_F (y|x)$. After introducing the prompter model $\rho$, the probability of generating response $y$ given input $x$ (denoted by $\tilde \pi_\rho$) can be expressed as:
\begin{equation}
    \tilde \pi_\rho(y|x)=\sum_{x'} \pi_F(y|x')\rho(x'|x).
\end{equation}
which captures the probability of generating the response $y$ for a given $x$ under the influence of the prompter $\rho$.

While conventional prompt optimization can be represented through probabilistic distributions \cite{trivedi2025align}, the new prompt optimization paradigm augmented with our objective optimization is formally expressed as:
\begin{equation}
    \max \limits_{\rho}E_{x'\sim \rho(·|x),y\sim \pi_F(·|x')}[Reward(\Delta I, \Delta H, \Delta SR)].
\end{equation}
Whereas the reward function in the initial prompt optimization formulation is denoted as $r^*(x,y)$, we reconfigure it into a tripartite reward structure $Reward(\Delta I, \Delta H, \Delta SR)$ based on our proposed desire architecture for LLM-based agents and goal function optimization procedures. This yields desire-biased modified prompts for subsequent decision-making processes.
\subsection{Step 3: Decision and Action Generation}
In societal simulations, the pre-trained language models employed by LLM-based agents remain static throughout operation. Given this constraint, direct model modification to accommodate optimized objective functions is infeasible. Instead, more precise emotion alignment responses are achieved via L2-generated modified prompts. Building upon the Reinforcement Learning from Human Feedback (RLHF) optimal decision formulation \cite{peng2019advantage}, we derive the following decision-generation equation for our framework:
\begin{equation}
    \pi^*(y|x)=\frac{1}{Z^*(x)}\pi_F(y|x)exp(\frac{Reward(\Delta I, \Delta H, \Delta SR)}{\beta}).
\end{equation}
where $Z^*(x)=\Sigma_y \pi_F(y|x)exp(Reward(\Delta I, \Delta H, \Delta SR)/\beta)$ is the normalizing constant, and $\beta$ is the alignment tuning parameter.

Furthermore, during the decision-making process, LLMs both output the decision result and generate a reasoning process (rationale) for agents to make future decisions, including what factors contributed to the decision and the relationship between the various factors. Too much knowledge can undermine the believability of LLM's simulation, as LLM-based agents may inadvertently express behaviors that is at odds with their state \cite{verspoor2024fighting}. The rationale, on the other hand, can further solidify the LLM-based agents' cognitive reasoning by showing the "knowledge-state" reasoning process to the LLM, which ensures the consistency among state, knowledge and decision.

In addition, Memory maintains a comprehensive record of agents' emotional decisions, including the problem that the agent encountered, the decision that LLM generated, and the reason for generating such decision.
When a new round of decision-making is executed, LLM-based agents need to make a reasonable decision based on the memory and the current state. First, it needs to judge whether the content in the memory structure is valid or not according to three indices. 1) Similarity. The emotional decision-making process of LLM-based agents is encoded into vector representations through \textit{all-MiniLM-L6-v2}, which are subsequently evaluated via cosine similarity metrics against predefined emotional categories. 2) Importance. This framework introduces a LLM-based evaluation agent to extract and analyze the generated rationales to assess emotional influence in decision-making. 3) Timeliness. When the time interval between the decision time and the creation of a memory is too long, LLM-based agents will remove the memory.

\section{Experimental Evaluation}

LLM-based agents are garnering increasing attention in the fields of artificial intelligence and social science due to their ability to simulate human-like responses and behaviors. The advantages of such social simulations include enhanced efficiency, reduced costs, improved scalability, and the circumvention of ethical concerns associated with human subjects—exemplified by the classic Stanford Prison Experiment \cite{zimbardo1971stanford}. However, significant disadvantages also exist, such as low transparency, poor reproducibility, and uncertainty regarding "simulation fidelity" \cite{horton2023large}. This section utilizes a food delivery scenario to simulate the interactions of key agent roles and the social emergence (involution) within it, thereby evaluating the proposed agent modeling framework. We carry out extensive experiments to answer the following research questions:
\begin{itemize}
\item \textbf{RQ1}: Can it achieve the social emergence that exhibits greater consistency with real-world social system? 
\item \textbf{RQ2}: Can it effectively capture the bounded rationality inherent to agents within the social system? 
\item \textbf{RQ3}: Do the agents exhibit greater consistency in their states, expectations, and behaviors?
\end{itemize}

\subsection{Experimental Setup}
 \textbf{Computational Resources.} Since we do not alter the parameters of the frozen model, our experiments require relatively fewer computational resources. Consequently, we were able to conduct all our experiments using a machine equipped with an Intel(R) Xeon(R) Gold 5418Y processor with two GeForce RTX 4090 GPUs. We used Python 3.11 to execute the experiments. 
 \textbf{Datasets}. We compared real-world data from the Zomato platform 2349 (including food delivery orders from multiple cities) with our multi-agent system \cite{dataDlivery}.

\textbf{Simulator}. This paper employs computational experiments \cite{xiao2023putational, xue2023computational, xue2024computational-1} and generative explanation methods \cite{xue2024computational, 10292987} to conduct an O2O delivery service ecosystem to evaluate the proposed emotion alignment framework \cite{xue2021computational, lu2021computational}. The virtual delivery scenario consists of four types of agents: 
\begin{itemize}
    \item Delivery riders. They are responsible for the delivery of takeaway orders, and are rewarded for completing the orders. This type of agents are the main research subjects in our experiments, constructed by different frameworks.
    \item Delivery service platform. It is responsible for the distribution of orders and sets up a list of income for the delivery personnel who participate in the work.
    \item Delivery bookers \& Dlivery makers. As NPCs in this social scene, they only need to complete the established process tasks and do not have the ability to adjust their behaviors.
\end{itemize}

Experimental scenario contains six rider Agents working in the virtual environment for a total of 30 days. The initial speed of each rider is set to 80, while The order acceptance probability is stratified into three discrete tiers: 30\%, 60\%, and 90\%. When the rider Agent starts to work, it needs to walk to the target workplace to wait for the order; after receiving the assigned order, it needs to judge whether to accept the order according to its own status and the order's status; if accept, it goes to pick up the order at the restaurant according to the route given by the platform and delivers the order to the service consumers (note that the maximum number of orders held simultaneously is 3); if reject and not currently held, it can continue to wait or randomly wander (change the waiting place); when it's time to sleep, the agent must complete the order in hand before rest.

\textbf{Experimental Design}.
This paper conducts two simulation experiment sets:

1) Single Agent-Type Simulations:
Five distinct agent architectures—rule-driven decision-making, imitation learning-driven decision-making, reinforcement learning-driven decision-making, GPT-4o-driven decision-making, and our framework-driven decision-making—were deployed in identical experimental setups. Key metrics including agent state trajectories, decision quality, and behavioral consistency were recorded.

2) Mixed Agent-Type Simulations:
Three architectures (reinforcement learning-driven decision-making, GPT-4o-driven decision-making, and our framework-driven decision-making) were co-simulated within a shared environment across three experimental runs. Cross-architecture metrics (comparative state dynamics, decision quality, and inter-agent consistency) were documented.

Simulation results were benchmarked against real-world datasets. Due to computational constraints, exact parametric alignment with empirical data was unattainable. Consequently, subsequent analysis prioritizes developmental trend congruence over point-to-point matching.

\textbf{Metric}. We quantify \textit{Involution} through the inverse of the coefficient of variation of agents’ incomes, denoted as Involution(t). A higher inverse coefficient of variation signifies more intense competition, thereby effectively capturing the dynamic changes in competition within the system: 
 \begin{equation}
  Involution(t)=\frac{\pi(t)*\theta(t)}{\mu(t)}
\end{equation}
where $\mu(t)$, $\theta(t)$ are the mean and standard deviation of riders' money at time $t$ and $\pi(t)$ is the mean of riders' walking distance at time $t$. This metric quantifies evolutionary trajectories within experimental societal systems, evaluating whether agent simulations of the current type demonstrate closer approximation to system evolution congruence observed in real-world datasets.

In addition, to quantify cognitive-behavioral consistency in LLM-based agents' state-goal-decision loops, we employ Dynamic Time Warping (DTW) \cite{yadav2018dynamic} to capture shape similarity across metric evolution curves. Crucially, agents demonstrating logically coherent and temporally consistent decision-making exhibit correlated patterns among objective optimization trends, trajectories of desire generation and state variation patterns. DTW identifies optimal alignment between temporal sequences, allowing different step sizes of the time axis and capturing the similarity between sequences, expressed as (recursive form):
\begin{equation}
\begin{aligned}
    DTW(i,j)=|x(i)-y(j)|+min(DTW(i-1,j), \\DTW(i,j-1), DTW(i-1,j-1)).
\end{aligned}
\end{equation}
\begin{figure}[h]
    \centering
        \subfigure[Involution from single type of agents]{\includegraphics[width=0.48\linewidth]{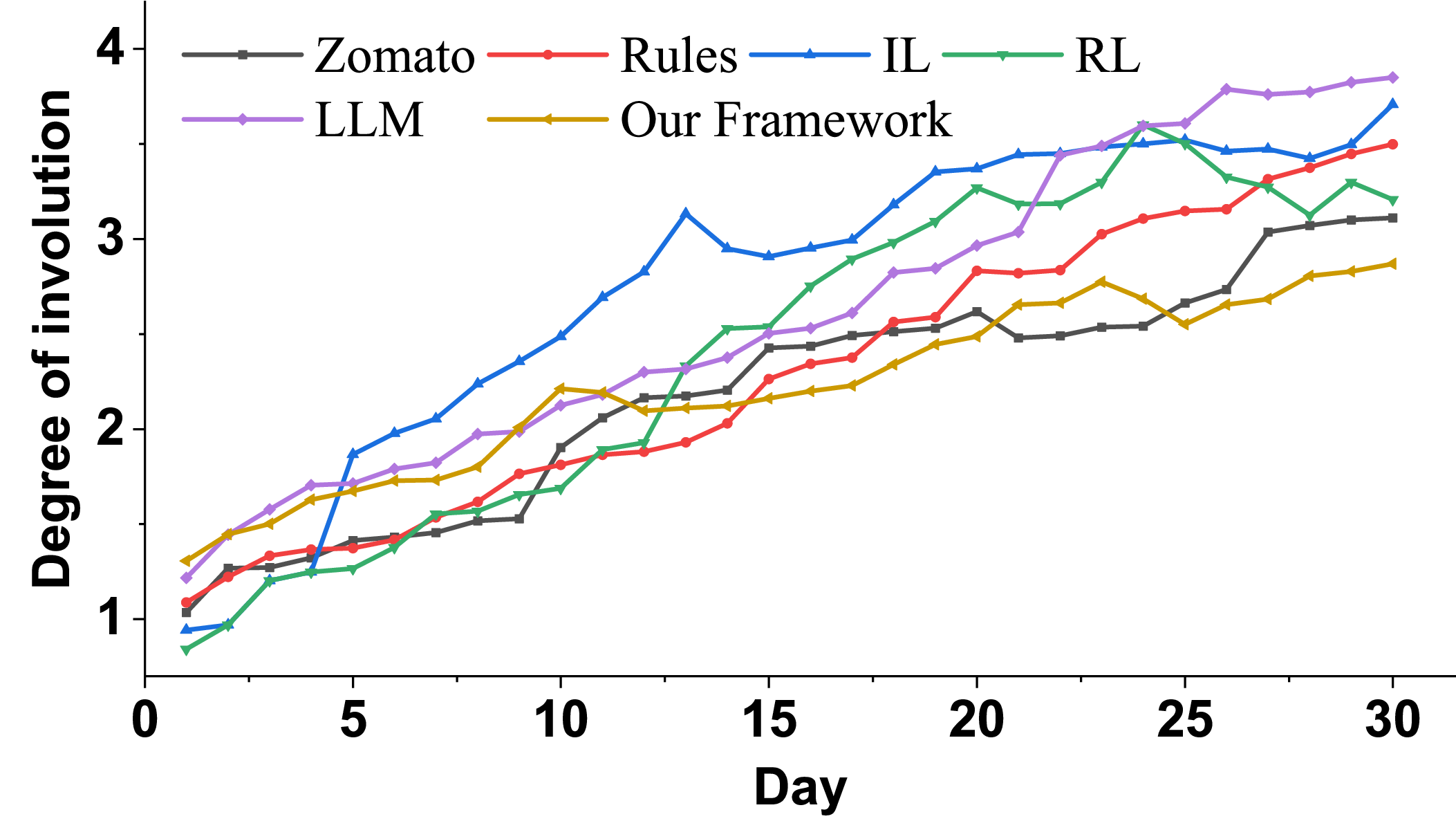}} 
        \subfigure[Involution and Performance from three mixed types of agents]{\includegraphics[width=0.48\linewidth]{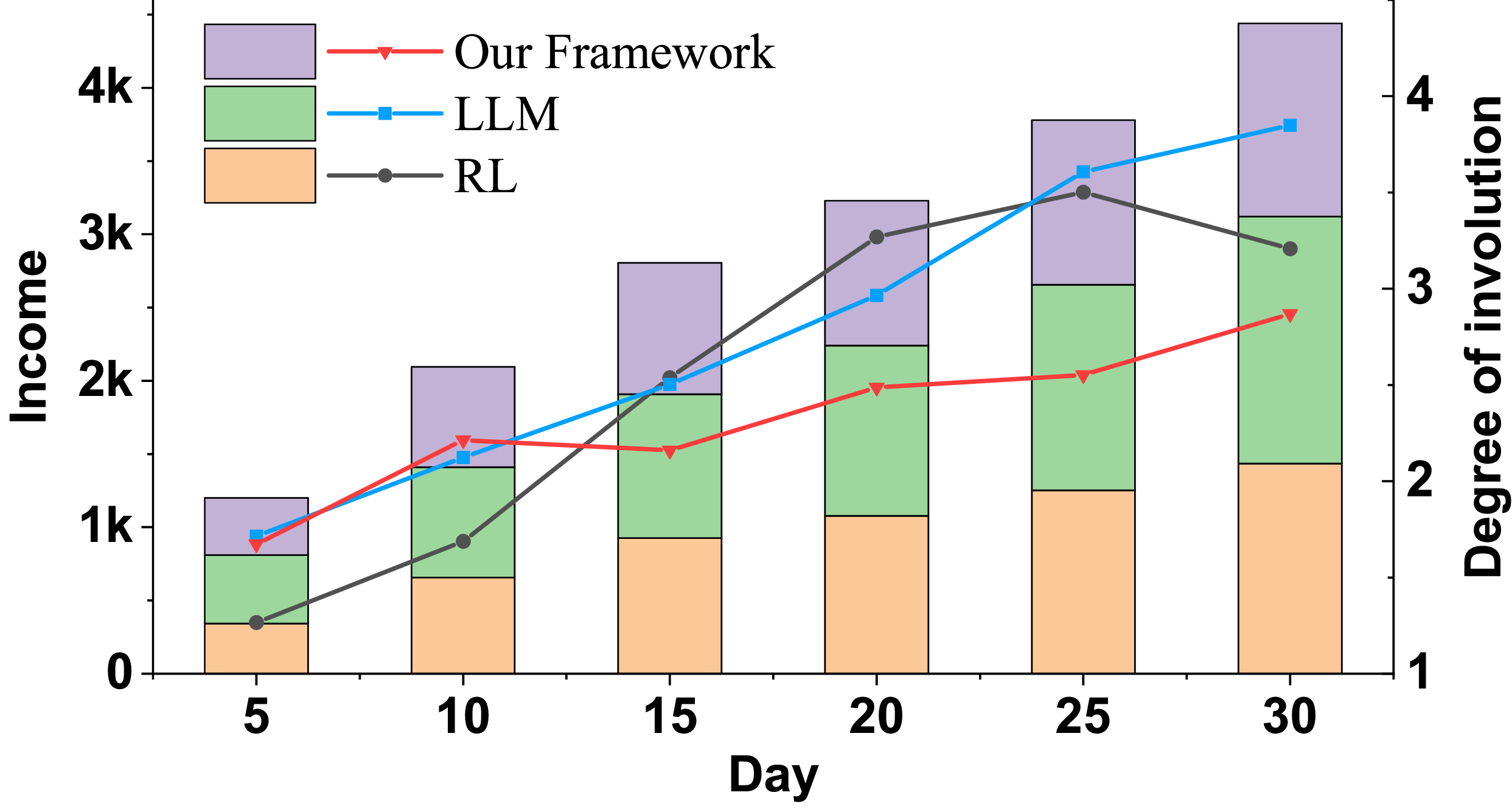}}\\ 
           \caption{Overall performance of different Rider Agents in O2O experimental scenario.}
    \label{wildlife}
\end{figure}
\subsection{RQ1-Social Emrgence: Degree of Involution}
We implemented five distinct decision frameworks—including our proposed framework—to model rider agents in the multi-agent interaction system: 1) Rule-based decision-making, 2) Imitation learning-driven, 3) Reinforcement learning-driven, 4) GPT-4o-driven, 5) Our affective cognition framework-driven. Experimental metrics included agent Income and Walking Distance for quantifying system involution levels. Figure 3(a) compares involution metrics between agent societies and real-world datasets, demonstrating that our affective cognitive modeling framework outperformed alternatives in generating emergence phenomena congruent with real social systems. Figure 3(b) presents results from mixed societies containing reinforcement learning-driven, GPT-4o-driven, and our framework-driven agents. Agents implementing our framework exhibited enhanced behavioral stability and superior decisional consistency amidst competitive heterogeneous interactions.

\begin{figure}[h]
    \centering
     \subfigure[Decision from single agents of our framework]{\includegraphics[width=0.34\linewidth]{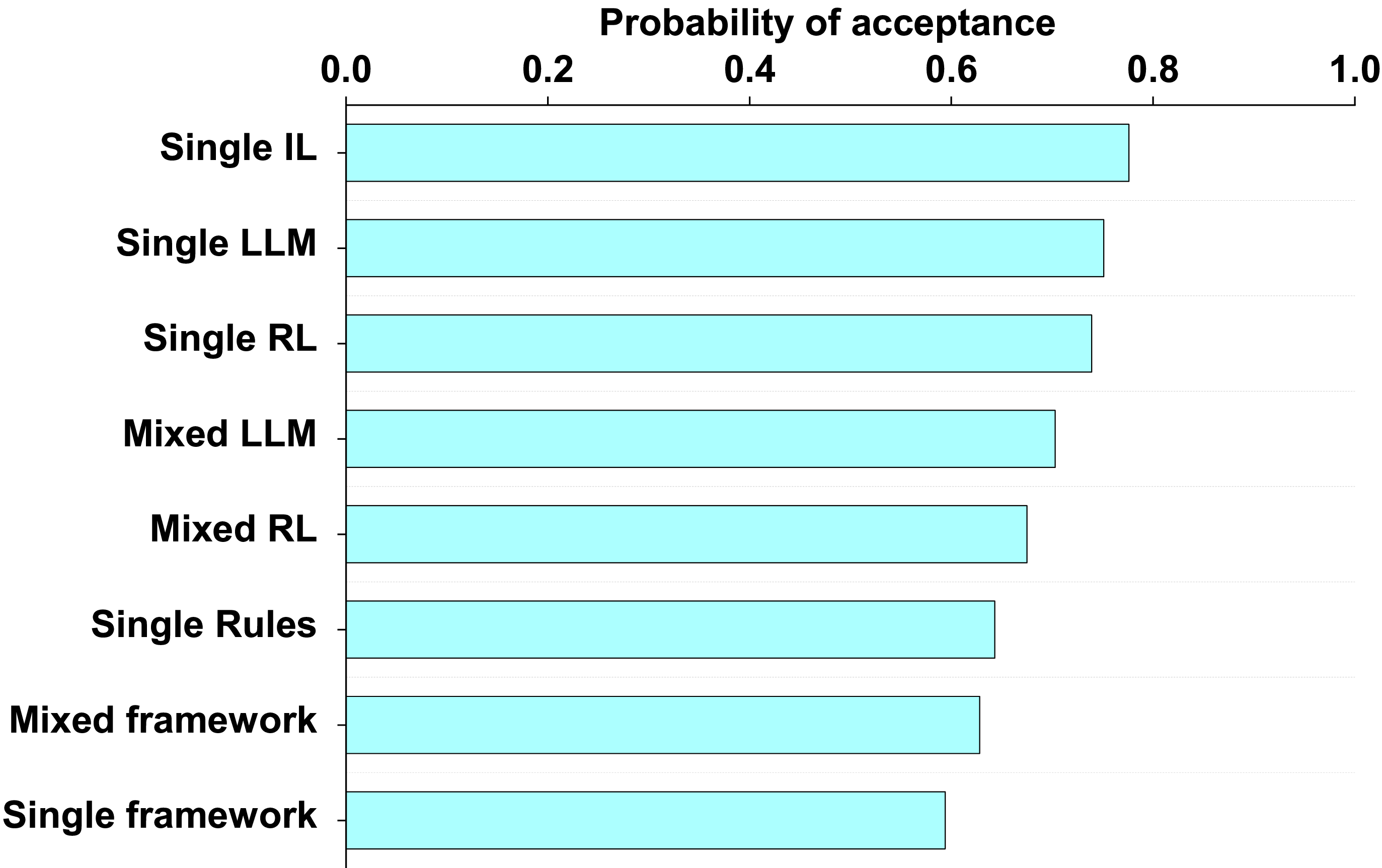}}
        \subfigure[Desire from single agents of our framework]{\includegraphics[width=0.3\linewidth]{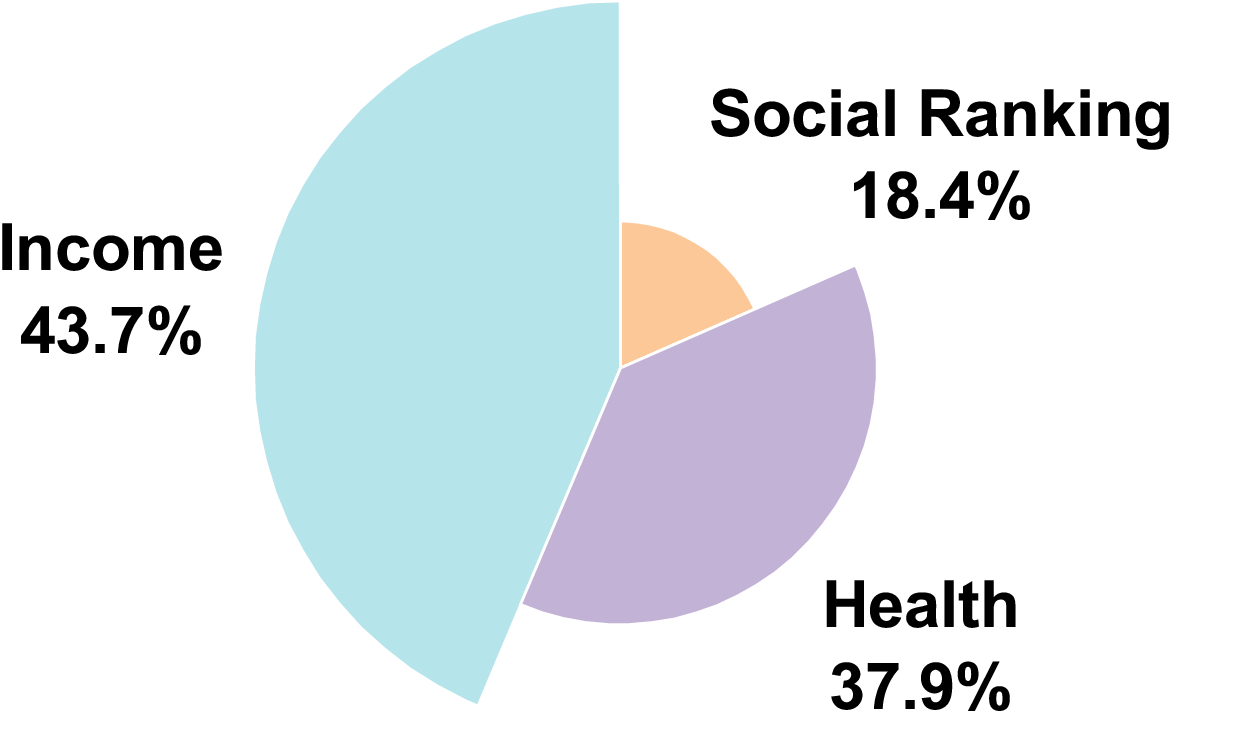}} 
        \subfigure[Desire from mixed agents of our framework]{\includegraphics[width=0.3\linewidth]{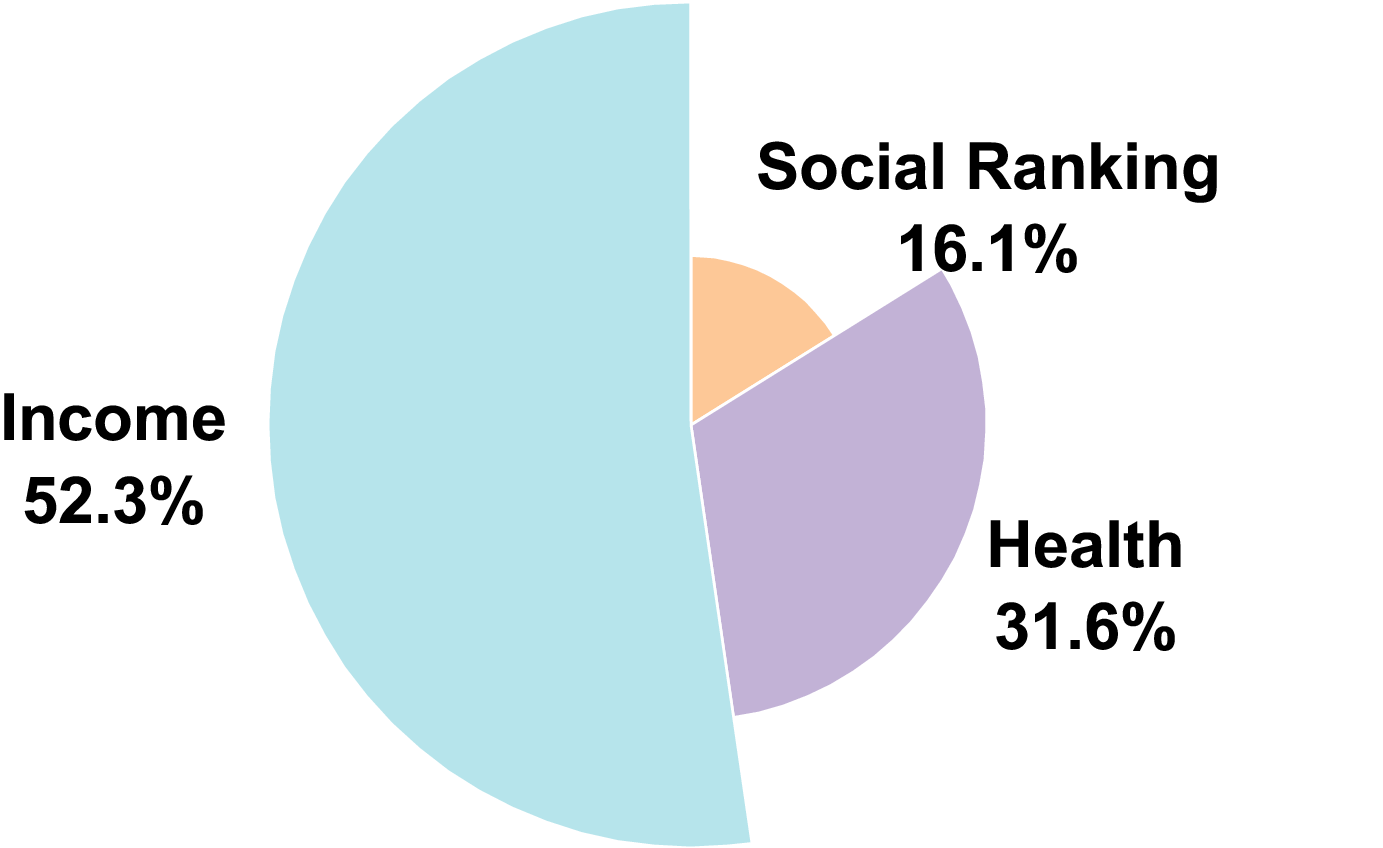}}\\ 
           \caption{Decision \& desire percentage of different Rider Agents in O2O experimental scenario.}
    \label{wildlife}
\end{figure}
\subsection{RQ2-Bounded Rtionality: Decision and Desire Percentage}
Following 30-day simulations, agents implementing our framework demonstrated significantly lower order acceptance rates compared to alternative architectures, as evidenced in Figure 4(a). This divergence stems from competing agents' over-optimization of income-maximization objectives, where health status and operational costs received insufficient weighting in their decision functions. Consequently, these agents uncritically prioritized high-yield orders—including economically inefficient low-quality high-yield orders, while disregarding health deterioration and task execution costs. Figures 4(b) and 4(c) visualize the desire distribution trajectories of our framework-driven agents across the simulation period.

\subsection{RQ3-Consistency: Similarity between Sequences}
\begin{figure*}[t!]
    \centering
        \subfigure[RL Agent]{\includegraphics[width=0.32\textwidth]{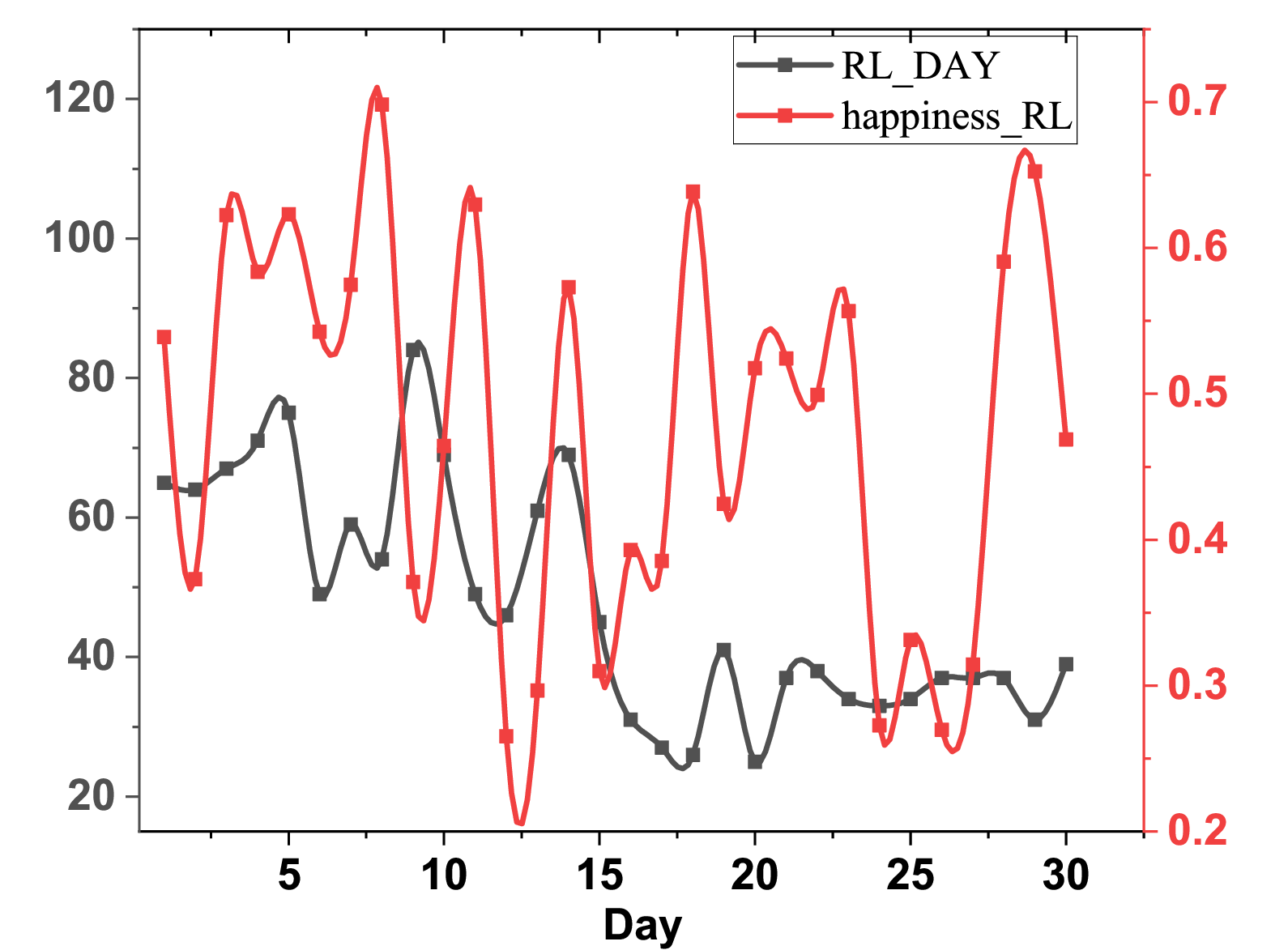}} 
        \subfigure[GPT-4o Agent]{\includegraphics[width=0.32\textwidth]{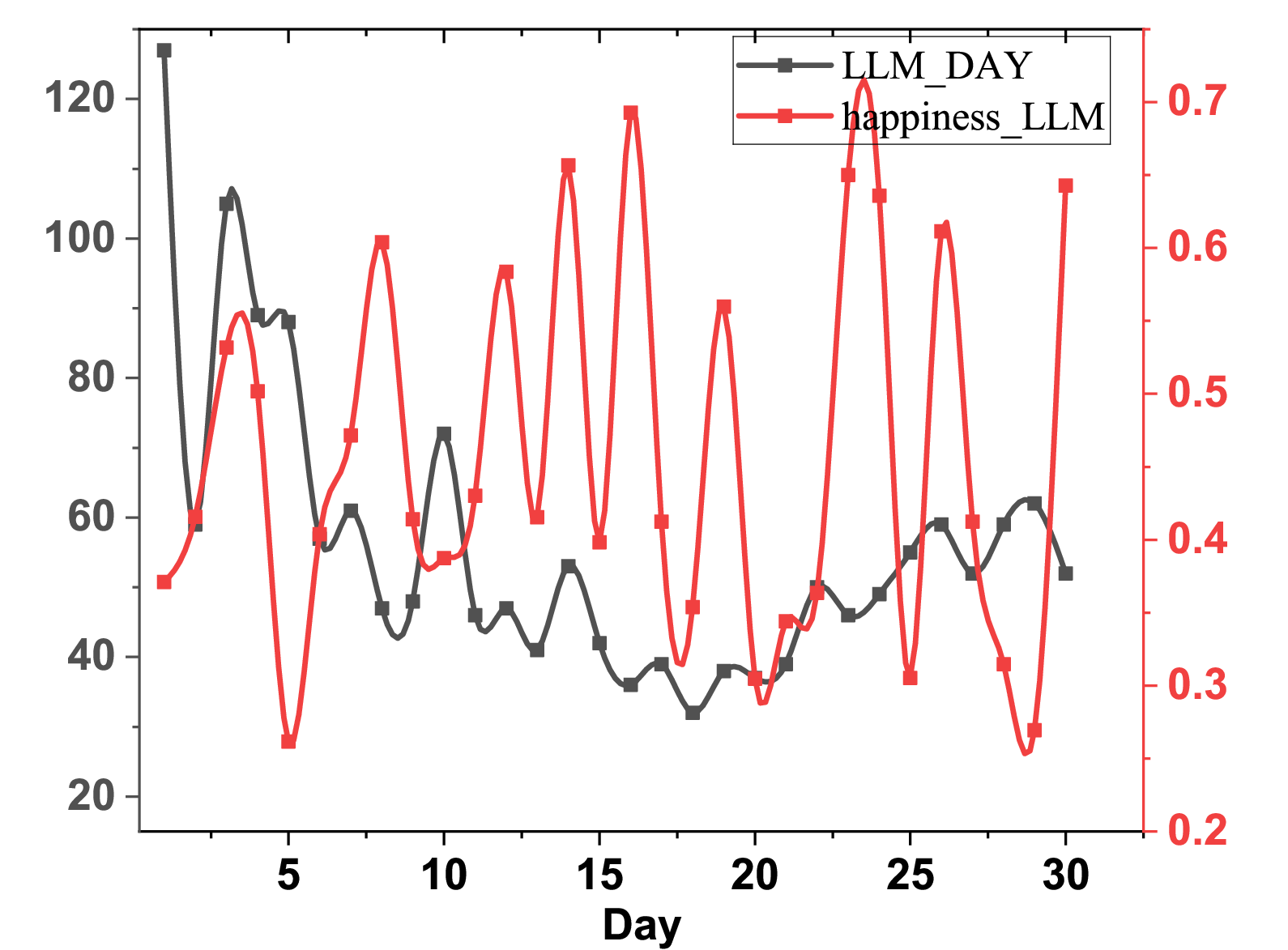}} 
        \subfigure[Our framework Agent]{\includegraphics[width=0.32\textwidth]{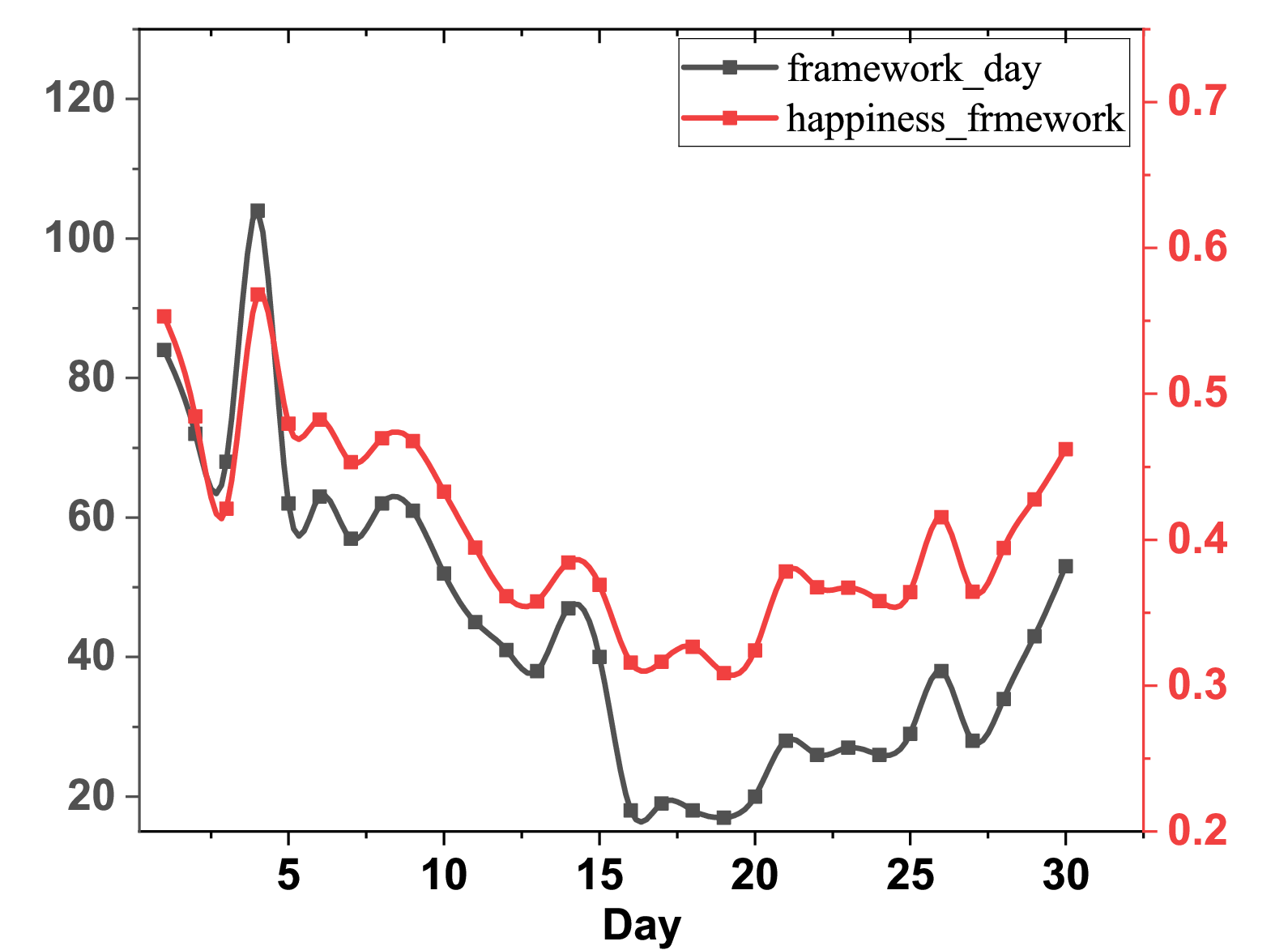}}\\         
           \caption{Money per day and happiness percentage per day of different Rider Agents in O2O experimental scenario.}
    \label{wildlife}
\end{figure*}

To investigate "State-Desire-Behavior" coherence in agents during simulation, this study employs Dynamic Time Warping (DTW) to assess similarity between daily income trajectories and daily "happiness"  variations across agent types, where income dynamics reflect behavioral impacts following desire activation, goal optimization, and decision execution, while "happiness" fluctuations capture affective responses to state changes that trigger new desires. Theoretically, material state changes (e.g., income) modulate affective states, motivating decisions to alter existing conditions. Figure 5 reveals that in mixed simulations: reinforcement learning (RL) agents (Fig 5a, DTW=274.24) and GPT-4o-driven agents (Fig 5b, DTW=326.29) exhibit pronounced affective volatility yet maintain stable daily income, indicating algorithmically constrained rational decision-making with decoupled affective-behavioral dynamics and absence of coherence; conversely, our framework agents (Fig 5c, DTW=265.08) demonstrate greater affective stability with synchronized income-affective trajectories, where negative-affect-driven income slumps during declining mood states manifest behavioral anomalies causing income loss, thereby evidencing emergent State-Desire-Behavior coherence of our framework agents through statistically DTW differentials (9.16 vs RL, 61.21 vs GPT-4o).

\subsection{Performance on Different LLMs}
\textbf{Datasets and Setup.} We evaluate the proposed framework on 3 distinct generative tasks: Dialogue Response Generation, Sentiment Reversal, and Constrained Sentence Generation. We instantiated our method and applied it to different BaseLLMs to accomplish generative tasks. Our primary objective was to evaluate whether our framework could be employed to enhance the performance of arbitrary BaseLLMs. To this end, we conducted comparative experiments between the enhanced LLMs and their original counterparts. Several prominent BaseLLMs were systematically employed across all experimental tasks: ChatGPT (gpt-3.5-turbo), DeepSeek-V3, Qwen2.5 and Gemma3. 

\textbf{Results.} Given the absence of established metrics applicable to our selected tri-category task datasets, we implemented a dual evaluation protocol. The first approach involved human assessment of model outputs: this evaluation was conducted by multiple authors through a controlled process where one author provided instructional inputs while the remaining evaluators independently selected outputs that best aligned with task requirements from responses generated by different models. The second method employed DeepSeek for task evaluation, utilizing structured prompting templates to systematically identify responses demonstrating superior instruction-task consistency across model outputs. The experimental results, as presented in Table I, reveal two critical observations: (1) LLMs equipped with COGNITIVE-enhanced mechanisms demonstrated superiority responses over BaseLLMs across all tasks, and (2) DeepSeek*, ChatGPT*, Qwen2.5* and Gemma3* exhibited substantially strong capability manifestations. These findings collectively demonstrate that our cognitive modeling framework effectively augments generative performance in LLMs. Additionally, when processing diverse tasks, existing LLMs demonstrate superior generative response quality in contextually-prompted sentence generation (Sentence Gen) and dialogue generation (Dialogue Gen) tasks.  However, their performance significantly degrades on the Yelp sentiment analysis dataset, a domain requiring specialized emotional knowledge. The performance improvements enabled by our cognitive modeling framework highlight the importance of LLM-based agents alignment in enhancing simulation fidelity.

 \begin{table*}[t]
  \scriptsize
\setlength{\tabcolsep}{7.5pt}
\renewcommand{\arraystretch}{1.5}
\centering
\caption{Performance of the framework on various tasks using different base LLMs.}
\label{tablename}
\vspace{5pt}
\begin{tabular*}{\linewidth}{cp{0.81cm}p{0.81cm}p{0.81cm}p{0.81cm}p{0.81cm}p{0.81cm}p{0.81cm}p{0.81cm}}\hline
\multirow{2}{*}{\textbf{Task}} &\multicolumn{2}{c}{ChatGPT} &\multicolumn{2}{c}{DeepSeek-V3} &\multicolumn{2}{c}{Qwen2.5} &\multicolumn{2}{c}{Gemma3}\\
\cline{2-9}   
& Base & +SE & Base & +SE & Base & +SE & Base & +SE\\ 
\cline{1-9} 
\textbf{Yelp} &9.2 &\textbf{41.5} &9.5 &\textbf{38.9} &9.1 &\textbf{37.2} &9.6 &\textbf{39.3}\\
\textbf{Dialogue Gen} &39.5 &\textbf{62.8} &42.1 &\textbf{69.2} &40.4 &\textbf{65.7} &44.8 &\textbf{68.5}\\
\textbf{Sentence Gen} &45.7 &\textbf{64.5} &44.6 &\textbf{63.1} &43.5 &\textbf{64.2} &46.9 &\textbf{65.4}\\
\hline
\end{tabular*}
\end{table*}

\section{Conclusion}
This study addressed a critical gap in social simulations by proposing a novel emotional cognition framework for LLM-based agents. We identified that the limited affective cognition and lack of bounded rationality in existing agents hinder the ecological validity of virtual social systems. To bridge this gap, our framework formally integrated desire generation and objective management to model a complete, emotion-driven decision-making process, achieving a closer alignment with human psychology.

The experimental implementation within our multi-agent environment yielded two key findings. First, agents governed by our framework consistently exhibited behaviors that were congruent with their internal emotional states. Second, and more significantly, in comparative assessments, these agents demonstrated superior ecological validity, with their decision outcomes approximating human behavioral patterns more closely than those of other agent types. This demonstrates that explicitly modeling emotional cognition is not merely an additive feature but a fundamental component for creating realistic and trustworthy agent-based societies.

The primary contribution of this work is twofold: (1) a structured architecture that embeds empirically grounded emotions into the core decision-loop of LLM-based agents, and (2) empirical evidence that this approach enhances the realism of social emergence in simulations. These findings have important implications for developing more reliable testbeds for social science research, complex system modeling, and the eventual deployment of human-aligned AI in interactive services.

Despite its promising results, this work has limitations. Our framework was validated in a specific proprietary environment, and its generalizability across vastly different social contexts remains to be tested. Furthermore, the complexity of human emotions suggests that our model of desire and objective management is a significant but preliminary step.

Future work will focus on expanding the diversity and granularity of emotional states and social scenarios within the simulation environment. We also plan to explore the integration of long-term memory and personality traits to foster even more consistent and individualized agent behaviors. Ultimately, we believe this line of research is a crucial step toward creating virtual societies that can serve as robust platforms for understanding and predicting complex human social dynamics.

\section*{Declaration of competing interest}
 The authors declare that they have no known competing financial interests or personal relationships that could have appeared to influence the work reported in this paper.

 \section*{Data availability}
 Data will be made available on request.

\section*{Acknowledgements}
This work has been supported in part by National Natural Science Foundation of China (No. 62472306, No. 62441221, No. 62206116), Tianjin University's 2024 Special Project on Disciplinary Development (No. XKJS-2024-5-9), Tianjin University Talent Innovation Reward Program for Literature \& Science Graduate Student (C1-2022-010), and Henan Province Key Research and Development Program (No.251111210500).





 \bibliographystyle{elsarticle-num} 
 \bibliography{reference}



\end{document}